\documentclass[onecolumn,11pt]{article}
\usepackage[top=1in, bottom=1in, left=1in, right=1in]{geometry}

\usepackage{microtype}
\usepackage{graphicx}
\usepackage{subfigure}
\usepackage{booktabs} 

\usepackage{hyperref}

\usepackage{amsmath}
\usepackage{amssymb}
\usepackage{mathtools}
\usepackage{amsthm}
\usepackage[usenames,dvipsnames]{xcolor}

\usepackage[capitalize,noabbrev]{cleveref}

\usepackage{svg}
\usepackage{epstopdf}

\usepackage{grffile}
\graphicspath{{"/FIGS"}}

\newcommand\blfootnote[1]{%
  \begingroup
  \renewcommand\thefootnote{}\footnote{#1}%
  \addtocounter{footnote}{-1}%
  \endgroup
}

\newcommand{\norm}[1]{\left\lVert#1\right\rVert}

\newcommand{\mb}[1]{\mathbf{#1}}

\newcommand{\R}{\mathbb{R}}

\DeclareMathOperator*{\tr}{tr}

\usepackage[textsize=tiny]{todonotes}

\usepackage{ifthen}
\usepackage{enumitem}
\newcommand{\Outline}[4]{%
  
  \color{Sepia}
  \setlist[enumerate]{nosep}
  \vspace*{-\baselineskip+1mm}
  \hrulefill
  \vspace*{-\baselineskip+1mm}

  \textbf{Topic Sentence:}\\
  #1 \vspace*{-\baselineskip+3mm}

  \textbf{Main Points:}
  \vspace*{-\baselineskip+2mm}
  \begin{enumerate}[nosep]
    #2
  \end{enumerate}\vspace*{-\baselineskip+3mm}

  \def\temp{#3}
  \ifx\temp\empty
  \else 
    \textbf{Key Citations:}
    \vspace*{-\baselineskip+2mm}
    \begin{enumerate}[nosep]
      #3
    \end{enumerate}\vspace*{-\baselineskip+2mm}
  \fi

  \def\temp{#4}
  \ifx\temp\empty
  \else 
    \textbf{Figure:}
    \vspace*{-\baselineskip+2mm}
    \begin{enumerate}[nosep]
      #4
    \end{enumerate}\vspace*{-\baselineskip+2mm}
  \fi

  \vspace*{-\baselineskip+2mm}
  \hrulefill
  \color{Black}
  \setlist[enumerate]{}
}

\newcommand{\edit}[1]{
{#1}
}

\title{{\fontsize{16}{16}\selectfont \textbf{Estimating Dynamic Flow Features in Groups of Tracked Objects}}} 

\author{\normalsize{Tanner D. Harms$^{1*}$, Steven L. Brunton$^{2}$, Beverley J. McKeon$^{1,3}$}\\
\footnotesize{$^1$ California Institute of Technology, Graduate Aerospace Laboratories, Pasadena, CA 91106, United States} \\
\footnotesize{$^2$ University of Washington, Department of Mechanical Engineering, Seattle, WA 98195, United States} \\
\footnotesize{$^3$ Stanford University, Department of Mechanical Engineering, Palo Alto, CA 94305, United States\vspace{-.2in}}
}

\date{}

\begin{document}

\maketitle

\blfootnote{$^*$ Corresponding author (tharms@caltech.edu).}
\vspace{-.2in}

\begin{abstract}
Interpreting motion captured in image sequences is crucial for a wide range of computer vision applications.  Typical estimation approaches include optical flow (OF), which approximates the apparent motion instantaneously in a scene, and multiple object tracking (MOT), which tracks the motion of subjects over time.  Often, the motion of objects in a scene is governed by some underlying dynamical system which could be inferred by analyzing the motion of groups of objects.  Standard motion analyses, however, are not designed to intuit flow dynamics from trajectory data, making such measurements difficult in practice.  The goal of this work is to extend gradient-based dynamical systems analyses to real-world applications characterized by complex, feature-rich image sequences with imperfect tracers.  The tracer trajectories are tracked using deep vision networks and gradients are approximated using Lagrangian gradient regression (LGR), a tool designed to estimate spatial gradients from sparse data.  From gradients, dynamical features such as regions of coherent rotation and transport barriers are identified.  The proposed approach is affordably implemented and enables advanced studies including the motion analysis of two distinct object classes in a single image sequence.  Two examples of the method are presented on data sets for which standard gradient-based analyses do not apply.  
\end{abstract}

\section{Introduction}
\label{sec:introduction}

Understanding dynamical behavior in sequences of images is crucial for many applications within science, engineering, and policy.  Because the motion of objects in a scene is often governed by an underlying flow, it conveys information of the system's dynamics and can be used for modelling, estimation, and control.  A prominent example of this type of analysis comes from the world of experimental fluid mechanics, where image sequences of illuminated tracers supply the basis for estimating underlying vector fields~\cite{Raffel.Kompenhans_PIVbook_2018} and associated flow metrics~\cite{Haller_LCS_2015}.  However, technological barriers have inhibited the generalization of such analyses to arbitrary systems of dynamic objects.  The algorithms that are effective with carefully structured fluids experiments are not, for instance, effective when applied to videos of herds of sheep or of traffic at an interchange.  Now, however, it is no longer difficult to identify arbitrary objects in images.  Moreover, recent developments in the data-driven analysis of dynamical systems enable estimation of vector-field gradients from sparse data.  Synthesizing these advances to estimate dynamical features in crowds of arbitrary objects is the goal of this work; if it is successful, there could be impact across broad disciplines, including drone surveillance, microfluidics, herds and swarms, traffic flows, crowd analysis, geophysical flows, and more.  

In the vision community, estimation of dynamical behavior is often pursued through one of two primary approaches.  The first of these is optical flow (OF), which identifies the apparent motion from one image to the next in a sequence.  A variety of surveys have been published to categorize the optical flow algorithms that have proliferated in recent years (for instance, \cite{Fortun.Kervrann_OFSurvey_2015, Zhai.Kong_OFSurvey_2021}).  Traditional approaches such as Lucas-Kanade \cite{BakerMatthews_L-KSurvey_2004} and Horn-Schunk \cite{HornSchunk_HSOF_1981} are knowledge-driven in the sense that they do not need to be trained on data to be implemented.  In recent years, deep, or data-driven, OF models such as FlowNet \cite{Dosovitskiy.Brox_FlowNet_2015, Ilg.Brox_FlowNet2_2017}, PWC-net \cite{Sun.Kautz_PWC-Net_2018}, and RAFT \cite{TeedDeng_RAFT_2020} have seen significant improvement in accuracy and computation time over their knowledge-driven counterparts \cite{Zhai.Kong_OFSurvey_2021}.  Scene flow---the extension of optical flow to three dimensions using either disparity (depth) estimation in frame or 3D point clouds obtained via LiDAR---has also been heavily researched in recent years due to breadth of applications \cite{Zhai.Kong_OFSurvey_2021}.  Though OF methods have become adept at estimating motion in scene, they are not well equipped to estimate the dynamic behavior of specific groups.  OF does not discriminate objects in a scene, so any analysis of the motion is done on all motion within the scene.  Moreover, OF analysis is limited to instantaneous measurements of velocity that are sensitive to the motion of the observer.  Thus it is not possible using OF alone to objectively study the group behavior of entities over time.  

Rather than broadly estimating the motion of an entire scene, multiple object tracking (MOT), examines the movement of particular objects over time.  It can be used to estimate the velocity of specific subjects along their trajectories rather than the velocity field over the entire scene.  Many reviews have been written on the subject (for instance, \cite{Xu.Li_DeepMOTSurvey_2019, Luo.Kim_MOTReview_2021}).  In its most basic form, MOT consists of an object detection step and a target association step.  Object detection may be accomplished through the use of large vision models \cite{Voigtlaender.Leibe_MOTS_2019}, and association achieved using estimation filters like Kalman filters or particle filters, the Hungarian algorithm, or one of many other approaches \cite{Luo.Kim_MOTReview_2021}.  Recent years have seen tracking frameworks built around vision transformers \cite{Su.Luo_TransTrack_2020, Meinhardt.Feichtenhofer_Trackformer_2022}, graph models \cite{Liu.Yu_GraphGSM_2020, Quach.Luu_DyGlip_2021}, and attention \cite{Guo.Shen_GraphAttention_2021, Chu.Yu_OnlineMOTAttention_2017}, to name a few alternative approaches.  

Traditional MOT approaches are also limited as a tool for identifying dynamical patterns in crowd behavior.  MOT analyses are designed to give information about the behavior of individual entities, rather than interpret the underlying systems driving their motion.  A field of work related to this is crowd analysis \cite{Zhan.Xu_CrowdSurvey_2008, Sanchez.Herrera_DeepCrowdAnalysis_2020}, which seeks to understand people's behavior in crowds.  MOT is sometimes used for such studies \cite{cheriyadatRadke_DominantMotionInCrowds_2008}, which has seen recent interest as a result of social distancing protocols during the COVID-19 pandemic \cite{Verma.Kulkarni_COVIDCrowdDetection_2021, Pouw.Corbetta_CrowdManPhysicalDistancing_2020}.  Crowd analyses, however, are typically not directed towards understanding the dynamic flow patterns espoused by the studied objects, but rather towards identifying anomalies or perceiving the intent of the crowd \cite{Sanchez.Herrera_DeepCrowdAnalysis_2020}.  Understanding flow patterns in crowds is still largely unaddressed.

Interpreting the dynamical systems governing the motion of groups of tracked objects is, however, an important aspect of experimental analysis of fluids, where decades of research has been devoted to extracting physical meaning from sequences of illuminated particle images.  The most common method for measuring fluidic flow fields is particle image velocimetry (PIV) \cite{Raffel.Kompenhans_PIVbook_2018}, which characterizes velocity fields through windowed correlations between frames of artificial particles illuminated in the flow.  Recently, however, algorithmic improvements have made Lagrangian particle tracking (LPT) \cite{Schanz.Schroder_ShakeTheBox_2016, SchroderSchanz_PTVReview_2023} a popular approach to flow field measurement.  Instead of examining the aggregate motion of particles in a window, trajectories of individual particles are traced and flow fields interpolated.  \edit{While PIV and LPT are robust and effective for analyzing flow fields in highly controlled experiments, there remains a need for measurement technologies that leverage naturally occurring and imperfect tracers (in the sense that they may be sampled from a broad distribution) and for methods which can be implemented successful using non-specialist imaging equipment.  The current state-of-the-art methods in flow measurement require expensive instrumentation and the addition of artificial tracers in order to be viable.}

Identifying dominant flow structures from the LPT particle trajectories is accomplished through the  theory of Lagrangian coherent structures (LCS) \cite{Haller_LCS_2015, AllshousePeacock_LagrangianBasedMethods_2015}.  Analyses in this field are typically designed to be objective \cite{Haller_LCStextbook_2023, Gurtin.Anand_ContinuumMechanics_2010}, such that the relative motion of the observer to the flow does not influence the results.  Because of this property and others, LCS have been influential for many applications in fluids literature, including forecasting of pollution patterns \cite{OlascoagaHaller_ForecastingPollutionPatterns_2012, Nolan.Ross_app.bio_2020}, understanding biological feeding mechanisms \cite{PengDabiri_app.bio_2009}, characterizing heart flows \cite{Shadden.Gerbeau_app.bio_2010}, and engineering improved fluid systems \cite{Ahmed.Hanifatu_app.aero_2023}.   

\edit{Many LCS analyses are based on high-resolution computation of flow gradients~\cite{AllshousePeacock_LagrangianBasedMethods_2015}.  This can pose serious limitations when working with naturally observed systems, where representative tracers are often sparsely distributed.  Some methods approximate true gradient-based metrics using few or even single trajectories~\cite{Haller.Encinas-Bartos_Quasi-ObjectiveDiagnostics_2021, Encinos-Bartos.Haller_objectiveEddyViz_2022}.  Recently, a regression-based approach to gradient estimation, known as Lagrangian gradient regression (LGR), was introduced to approximate complete gradients from sparse trajectories (\cite{Harms.McKeon_LGR_2023}, under review).  LGR enables typical LCS analyses and velocity gradient analyses from tracer distribution densities that are reasonable for naturally observed systems.  In the original paper, the algorithm performance was demonstrated on numerical benchmark data with known tracer trajectories.}


\edit{This work leverages modern capabilities in computer vision in tandem with LGR to provide a framework for extracting dynamical information beyond velocity from groups of observed objects.  Specifically, large vision models are used to identify objects of interest and record their motion as a series of trajectories.  These are provided as inputs to the Lagrangian gradient regression (LGR) algorithm which enables reliable gradient estimation from sparse trajectories such as are characteristic of MOT studies.   Using the flow gradients, quantitative information of the flow including rotation and deformation rates for instantaneous and finite times can be computed.  While such gradient-based analyses have been useful in fluidic applications \cite{Haller_LCS_2015, AllshousePeacock_LagrangianBasedMethods_2015}, this work aims to extend their use to real-world scenarios where dynamical object classes and the scenes they inhabit can be visually complex and difficult to analyze by standard approaches.}

Specific attributes of the proposed method include:
\begin{itemize}
    \item The technique reveals physically interpretable dynamical structures such as transport barriers and regions of coherent rotation directly from groups of tracked objects. 
    \item The approach is general, and therefore it can be applied to any system that is characterized by the motion of multiple objects captured in sequences of images.  Moreover, the method is effective even when the spacing of observed tracers is sparse.  
    \item The three stages of the approach---detection, tracking, and sparse gradient estimation---are modular, which allows new developments in technology to be seamlessly integrated into the process.
    \item Because multiple classes may be detected in a sequence, dynamical features in groups of disparate object classes can be simultaneously identified.  
    \item The tool is built to enable objective analyses, for which the motion of the observer (drone, vehicle, etc.) will not alter the results. 
\end{itemize}

The rest of the paper is organized as follows:  \edit{Section \ref{sec:detection and tracking} discusses the modern vision tools which are implemented to identify the sets of object trajectories required by LGR. Then, section \ref{sec:estimating dynamical features from trajectories} provides the theoretical framework for LGR, which is used to identify dynamical features from the sparse trajectories which MOT detects}.  Because appropriate benchmarks are not available for this type of analysis, the tool is demonstrated on a case study using data from an experimental debris flow in section \ref{sec:case study: experimental debris flow} and on field data in section \ref{sec:field test: pond debris}, with discussion in section \ref{sec:discussion}.  

\section{Detection and Tracking}
\label{sec:detection and tracking}

Acquiring trajectories for flow structure computation is achieved through modern object detection and tracking techniques.  Using image sequences, trajectories may be observed in either $d=2$ or $d=3$ dimensions.  In the first scenario, a single camera is assumed to record motion on a plane that is approximately parallel to the sensor plane or where a homography can reasonably orient the camera as perpendicular to the plane of motion.  For 3-dimensional motion, multiple cameras are required and an additional triangulation step is necessary between detection and tracking.  For the remainder of this work, only planar flows observed with a single camera are considered.  

Deep vision models provide the framework for robust and general object identification.  Many models, including RCNN architectures \cite{Girschick_Fast-RCNN_2015, He.Girschick_Mask-RCNN_2017}, YOLO architectures \cite{Redmon.Farhadi_YOLO_2016, Jiang.Ma_YOLOReview_2022}, transformer architectures \cite{Carion.Zagoruyko_DETR_2020}, and more are sufficient.  It is necessary to represent each detection as a single point in the flow space.  In this work, the centroid of an identified mask is used, although other approaches can be considered based on the problem parameters.  Large vision models such as those previously mentioned also allow for flexibility regarding applications.  If unconventional objects must be detected, then transfer learning techniques \cite{Zhuang.He_TransferLearning_2020} can be applied to train a custom head on the network while preserving the features identified in the backbone.  Many cases may also involve dense crowds of objects or relatively small objects.  In such instances, it may be helpful to incorporate a sliding window approach to detections, which will be discussed in more depth with the case study in section \ref{sec:case study: experimental debris flow}.

Using whatever detection model is preferred, object tracking algorithms are used to stitch independent detections into trajectories.  Many such algorithms exist in the computer vision literature, including filters and deep implementations \cite{Bewley.Upcroft_SORT_2016, Wojke.Paulus_DeepSORT_2017} as well as in the fluids literature from the Lagrangian particle tracking community \cite{Schanz.Schroder_ShakeTheBox_2016, SchroderSchanz_PTVReview_2023}.  Whatever tracking algorithm is used, the quality of the trajectories in frame should be emphasized.  Trajectory lengths should be made as long as possible while in the frame.  As will be seen later, truncated trajectories are the most significant failure point of the LGR algorithm.  However, while many computer vision applications are concerned with re-identification of objects that leave the frame, that is not an issue here.  Any objects that exit and re-enter the scene can be viewed as entirely new entities without loss of performance.  

The quality of LGR analysis depends the characteristics of the observer and the properties of the flow.  In planar flows, optimal measurements will be made when the camera sensor is aligned parallel to the flow and lens distortion is minimized.  Also, the spacing of objects in the flow effects the resolvable scales of motion that can be interpreted from the trajectory data.  If objects are tightly packed, then smaller scales of motion can be discerned from the analysis---the spacing of particles effectively acts as a low-pass filter in the LGR algorithm.  Since small and crowded objects are more difficult to identify, there is thus a tradeoff between flow structure resolution (which improves with higher density and smaller objects) and detection quality (which improves with larger and distinct objects).

The stages of the analysis are defined in a modular way to allow for various implementations of detection and tracking to be easily interchanged.  This enables the rapidly evolving detection and tracking technologies to seamlessly integrate into the flow structure identification process.  Moreover, it may be that some models perform better on certain data types than others, so it is desirable to be able to quickly switch one for another.

\section{Estimating Dynamical Features from Trajectories}
\label{sec:estimating dynamical features from trajectories}

\edit{Lagrangian gradient regression (LGR) is a data-driven tool recently developed for the purpose of extracting flow features from groups of dynamical objects.  This work implements LGR as the third stage in the proposed algorithmic framework for analyzing flow structures.  The theoretical underpinnings of LGR are here introduced and sample metrics which it enables are defined. The original paper (\cite{Harms.McKeon_LGR_2023}, under review) contains a thorough description of the theory and algorithms behind LGR as well as demonstrations using controlled numerical data with known tracer trajectories.}

\subsection{Lagrangian Gradient Regression}

The theoretical framework of LGR begins by considering solutions to some differential equation
\begin{equation}
    \dfrac{d\mb{x}}{dt} \equiv \dot{\mb{x}} = v(\mb{x},t),
    \label{eq:diffeq}
\end{equation}
where $\mb{x} = \mb{x}(t)\in D$ is a vector-valued function of time $t\in\R$, and the velocity field $v:D\to D$ is a smooth function on the flow domain $D\subseteq \R^d$.  In systems that are observable by camera, dimensionality $d \in \{2,\,3 \}$.  The flow map generated by equation \ref{eq:diffeq} is defined as 
\begin{equation}
    \mb{F}_{t_0}^{t}:D\to D \quad \text{such that} \quad \mb{x}_0\mapsto\mb{F}_{t_0}^{t}(\mb{x}_0)
    \label{dq:flowmap}
\end{equation}
for given initial conditions $\mb{x}_0$ and $t_0$.  Observed trajectories are thus defined as $\mb{x}(t; \mb{x}_0, t_0) = \mb{F}_{t_0}^{t}(\mb{x}_0)$ for some finite $t$ \cite{GuckenheimerHolmes_NonlinearDynamicalBifurcation_1983, Shadden.Marsden_FTLEProperties_2005}. Henceforth, trajectories will be referred to with arguments implicitly understood.  

The Jacobian of the flow map,
\begin{equation}
    D\mb{F}_{t_0}^t = \mb{\nabla}_{\mb{x}_0}\mb{F}_{t_0}^t,
    \label{eq:flowmapjacobian}
\end{equation}
is of particular interest for estimating dynamical behavior in a flow.  It represents a linear operator mapping initial conditions to deformed positions at some future time $t$.  

Numerically approximating the Jacobian typically involves propagating perturbations $\mb{y}_0 = \mb{x}_0 + \Delta \mb{x}_0$ from $t_0$ to $t$ with $\Delta \mb{x}_0$ small and performing finite-differences. 
However, if $\Delta \mb{x}_0$ is not small and $\Delta t = t-t_0$ is large relative to the time-scales of the dynamical system $\mathcal{T}$, the linear approximation of the deformation is no longer valid and the Jacobian suffers computational error.  However, due to the smoothness of trajectories, if $\Delta t \ll \mathcal{T}$, the linear approximation $D\mb{F}_{t_0}^t$ is appropriate.  That is, the deformation at time $t$ is
\begin{equation}
    \Delta \mb{x} = \mb{F}_{t_0}^{t\to t_0}(\mb{y}_0) - \mb{F}_{t_0}^{t\to t_0}(\mb{x}_0) = D\mb{F}_{t_0}^{t\to t_0} \Delta \mb{x}_0
\end{equation}
for some $\mb{y}_0$ a perturbation.  

The instantaneous flow map Jacobian is linked to the velocity gradient of the flow at the initial conditions by
\begin{equation}
    \mb{\nabla}\mb{v} = D\dot{\mb{F}}_{t_0}^t \left(D\mb{F}_{t_0}^t\right)^{-1}
\end{equation}
\cite{Gurtin.Anand_ContinuumMechanics_2010}. Therefore, since $D\mb{F}_{t_0}^{t_0} = \mb{I}_d$ the identity matrix, the velocity gradient can be simply approximated from the short-time flow map Jacobian by
\begin{equation}
    \nabla \mb{v} \approx \frac{D\mb{F}_{t_0}^{t_0 + \Delta t} - \mb{I}_d}{\Delta t}
    \label{eq:gradv_approx}
\end{equation}
for $\Delta t = t-t_0 \ll \mathcal{T}$.

Another result of trajectory continuity is that the product of successive short-time Jacobian computations yields the Jacobian over the full analyzed time domain.  Thus,
\begin{equation}
    D\mb{F}_{t_0}^{t_n}(\mb{x}_0) = \prod_{i=0}^{n-1}D\mb{F}_{t_i}^{t_{i+1}}(\mb{x}(t_i)),
    \label{eq:compeq}
\end{equation}
where the index $i$ iterates connected observations of the flow map Jacobian along trajectory $\mb{x}(t; \mb{x}_0, t_0)$ \cite{BruntonRowley_FastComputationFTLE_2010}.  

In observed flows, systems of identified trajectories can be used to fit flow map Jacobians through kernel-weighted least squares regression
\begin{equation}
    D\mb{F}_{t_0}^t \approx \mb{X}_t \mb{KX}_{t_0}^{\top} \left( \mb{X}_{t_0} \mb{KX}_{t_0}^{\top} + \gamma n \mb{I}_d \right)^{-1},
    \label{eq:kernelweightedregression}
\end{equation}
where $\mb{X}_{t_0},\, \mb{X}_{t} \in \R^{d\times n}$ record the distances from all perturbations $\mb{y}_j(t_i),\, j\in \{1,\,\dots,\,n\}$ to the analyzed position $\mb{x}(t_i)$ at times $t_i = t_0$ and $t_i = t$ respectively, $\mb{K}\in \R^{n\times n}$ is a symmetric positive definite kernel weighting matrix, and $\gamma$ is a regularizing constant.  

\edit{The gradients $\nabla \mb{v}$ and $D\mb{F}_{t_0}^t$ form the basis for analyzing the kinematic behavior of the flow and for identifying flow features.}

\subsection{Objective Metrics of Interest}
In situations where the frame of reference between the system and the observer is not well defined, it is important that metrics on the data are objective; that is, that they are unchanged by transformations of the type
\begin{equation}
    \tilde{\mb{x}} = \mb{Qx}+\mb{p},
\end{equation}
where $\mb{Q}$ is a proper orthogonal rotation tensor and $\mb{p}$ is a translation \cite{Haller_LCStextbook_2023}.  Objective metrics are especially important to modern applications of computer vision due to the likelihood that the observer's position is not fixed or not well known.  For instance, measurements taken by drone will be confused by the motion of the drone itself if the metrics are not objective.  However, it is important to note that, for planar observations with a single camera, $\mb{Q}$ and $\mb{p}$ do not account for out-of-plane rotation or translation, which can still corrupt typical objective measurements. Below, some examples of relevant objective metrics are presented.  

Both instantaneous and finite-time metrics can be constructed using LGR.  Examples of objective instantaneous metrics are the principal strain rate and the vorticity deviation.  To define these, the Euler-Stokes decomposition is applied to the velocity gradient
\begin{equation}
    \nabla \mb{v} = \mb{W} + \mb{D},
\end{equation}
separating it into a skew-symmetric spin tensor $\mb{W}=\frac{1}{2}(\nabla \mb{v} - \nabla \mb{v}^\top)$ and a symmetric stretch tensor $\mb{D} = \frac{1}{2}(\nabla \mb{v} + \nabla \mb{v}^\top)$ \cite{Gurtin.Anand_ContinuumMechanics_2010}.  Vorticity is defined as the unique vector $\mb{\omega}$ satisfying
\begin{equation}
    \mb{We}= - \frac{1}{2}\mb{\omega}\times \mb{e}, \quad \forall \mb{e}\in \R^d.
    \label{eq:vort}
\end{equation}
Vorticity itself is not objective, but vorticity deviation is \cite{Haller.Huhn_LAVD_2016}.  It is defined as 
\begin{equation}
    \mb{\omega}'(\mb{x},t) = \left\lvert\mb{\omega}(\mb{x},t) - \overline{\mb{\omega}}(t)\right\rvert,
    \label{eq:vortdev}
\end{equation}
where the overbar signifies spatial averaging.  The principal strain rate is given as maximum eigenvalue of the stretch tensor
\begin{equation}
    \epsilon_1 = \lambda_{max}(\mb{D})
    \label{eq:epsilon}
\end{equation}
and represents the magnitude of strain rate in the direction of greatest stretching.  Because the stretch tensor is already objective, $\epsilon_1$ is as well.  

Finite-time metrics require extended trajectory history for computation; prominent examples include the finite-time Lyapunov exponent (FTLE) \cite{HallerYuan_LCSAndMixing_2000, Shadden.Marsden_FTLEProperties_2005} and the Langrian-averaged vorticity deviation (LAVD) \cite{Haller_DynamicPolarDecompostion_2016, Haller.Huhn_LAVD_2016}.  These can be viewed as the finite-time representations of the principal strain rate and the vorticity deviation respectively.  The FTLE is defined as 
\begin{equation}
    \sigma_{t_0}^t = \frac{1}{\Delta t}\ln \left( \norm{D\mb{F}_{t_0}^t}_2 \right),
    \label{eq:ftle}
\end{equation}
and represents the exponential growth rate of a linear deformation from time $t_0$ to $t$.  FTLE analyses are commonly done in geophysical or fluid mechanical contexts where the deformation is volume preserving (i.e. $\frac{1}{d}\tr(D\mb{F}_{t_0}^t)=1$) and thus $\sigma_{t_0}^t \geq 0$.  In general, however, $\frac{1}{d}\tr(D\mb{F}_{t_0}^t)\neq1$.  Thus, in computer vision applications, $\sigma_{t_0}^t$ may be less than zero if observed objects are locally contracting.  The LAVD, on the other hand, gives twice the rotation observed over an object's trajectory, and is therefore always greater than or equal to zero \cite{Haller.Huhn_LAVD_2016}.  It is defined
\begin{equation}
    \text{LAVD}_{t_0}^t = \int_{t_0}^t \left \lvert \mb{\omega}(\mb{x},\tau) - \overline{\mb{\omega}}(\tau)\right\rvert d\tau = 2\psi_{t_0}^t,
\end{equation}
where the intrinsic rotation angle (IRA) $\psi_{t_0}^t$ is the amount of rotation observed over the duration.  

\section{Laboratory Test Case}
\label{sec:case study: experimental debris flow}

Here an illustrative example is used to demonstrate the performance and highlight features of the proposed method.  The chosen case study is of an experimental debris flow, where unconventional particles of various geometry are added to the flow for visualization.  
\begin{figure}[t]
    \centering
    \includegraphics[width = 0.6\columnwidth]{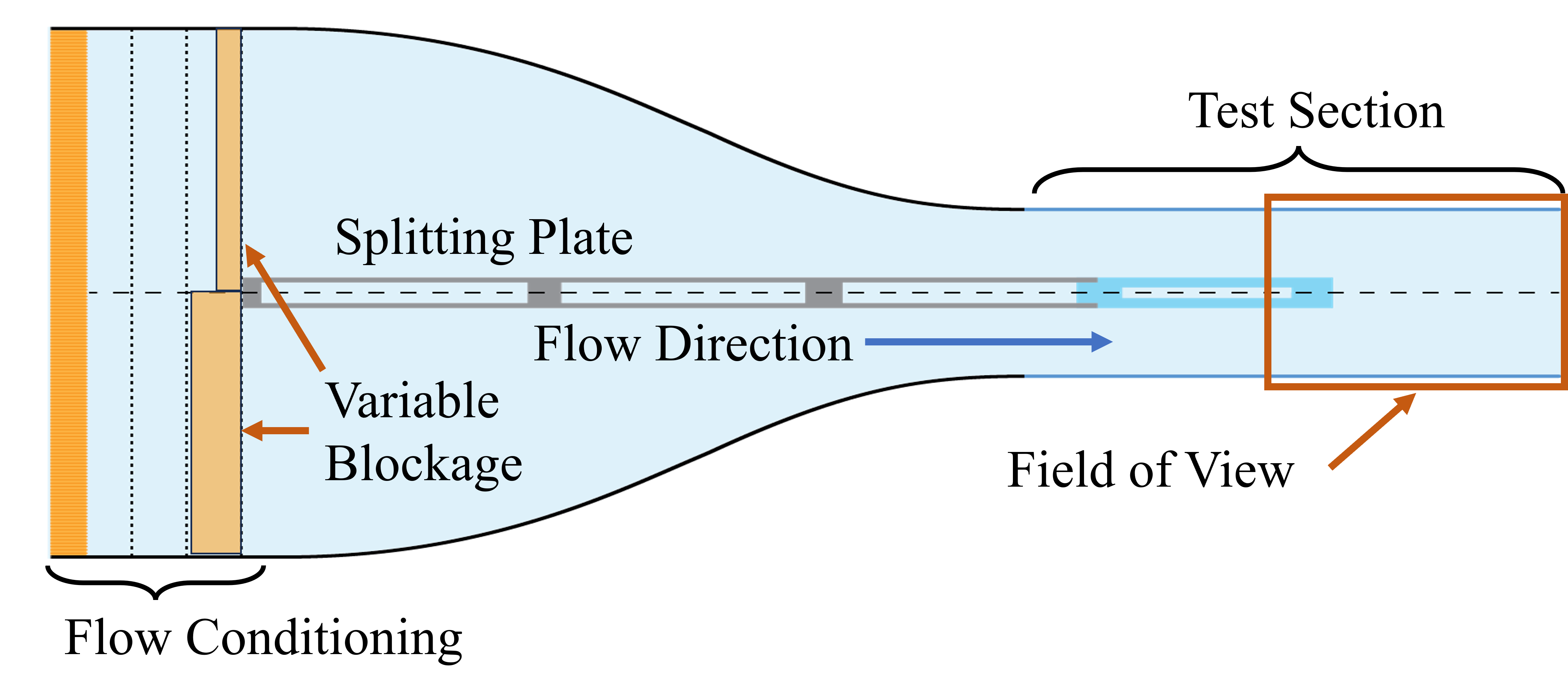}
    \caption{Schematic of the experiment used in the case study.}
    \label{fig:schematic}
\end{figure}

\subsection{Experimental Setup}
The experiment considered here as a demonstration was performed in the NOAH water channel at the Graduate Aerospace Laboratories at California Institute of Technology (GALCIT).  A $0.057$ meter thick plate was placed into the channel so that the trailing edge was situated near the streamwise midpoint of the $0.45 \times 0.45 \times 1.22$ meter test section.  A velocity ratio of $U_2/U_1 < 1$ was enforced by adding various blockage implements upstream.  The result is a flow at the trailing edge which exhibits a complex behavior consisting of bluff body vortex shedding that transitions to a shear layer. 

Generally accessible optics were used in this experiment.  While a machine vision camera was used in this case, the study could have been just as easily conducted with cell phone cameras. Specifically, video recording was performed using a FLIR Blackfly S BFS-U3-23S3C-C USB3 color camera with an Edmund Optics 4mm C-mount lens.  The optical setup was mounted directly above the water surface just downstream of the plate trailing edge and $1920\times 1200$ pixel images were recorded at 60Hz during the tests.  

\edit{
The optical system was calibrated using a ChArUco calibration plate.  The ChArUco calibration approach improves upon standard checkerboard calibration by filling the white space with ArUco fiducial markers, enabling robust calibration in circumstances where the entire checkerboard may not be visible.  For this experiment, the calibration plate was defined on a $28\times 17$ square checkerboard where each square edge was $0.02$ meters long and the contained ArUco markers were $0.015$ meters on each edge and displayed unique $5\times 5$ pixel patterns.  The calibration process involved recording many images ($\mathcal{O}(50)$) with arbitrary orientation of the board to the camera, which were used to calibrate the camera intrinsic matrix for the optical setup.  Additional images were then collected with the ChArUco board in the plane of the flow.  These served to fit a homography matrix which was used transform the camera pose to one situated directly perpendicular to the flow plane with the dimensional units of the flow.
}

To visualize the flow motion, unconventional tracers were added on either side of the splitting plate upstream of the trailing edge.  Three categories of tracers were considered: 1. $9.53$mm diameter birch spheres, 2. $19.05$mm diameter birch spheres, and 3. $6.35$mm diameter birch rods cut roughly between $12.7$ and $50.8$mm in length.  The following results will discuss two test cases:  In the first, only small birch spheres are used as flow indicators. The second uses all of the particle types simultaneously.  

\subsection{\edit{Gradient Estimation via Existing Methods}}
\begin{figure*}[t!]
    \centering
    \includegraphics[width=1\textwidth]{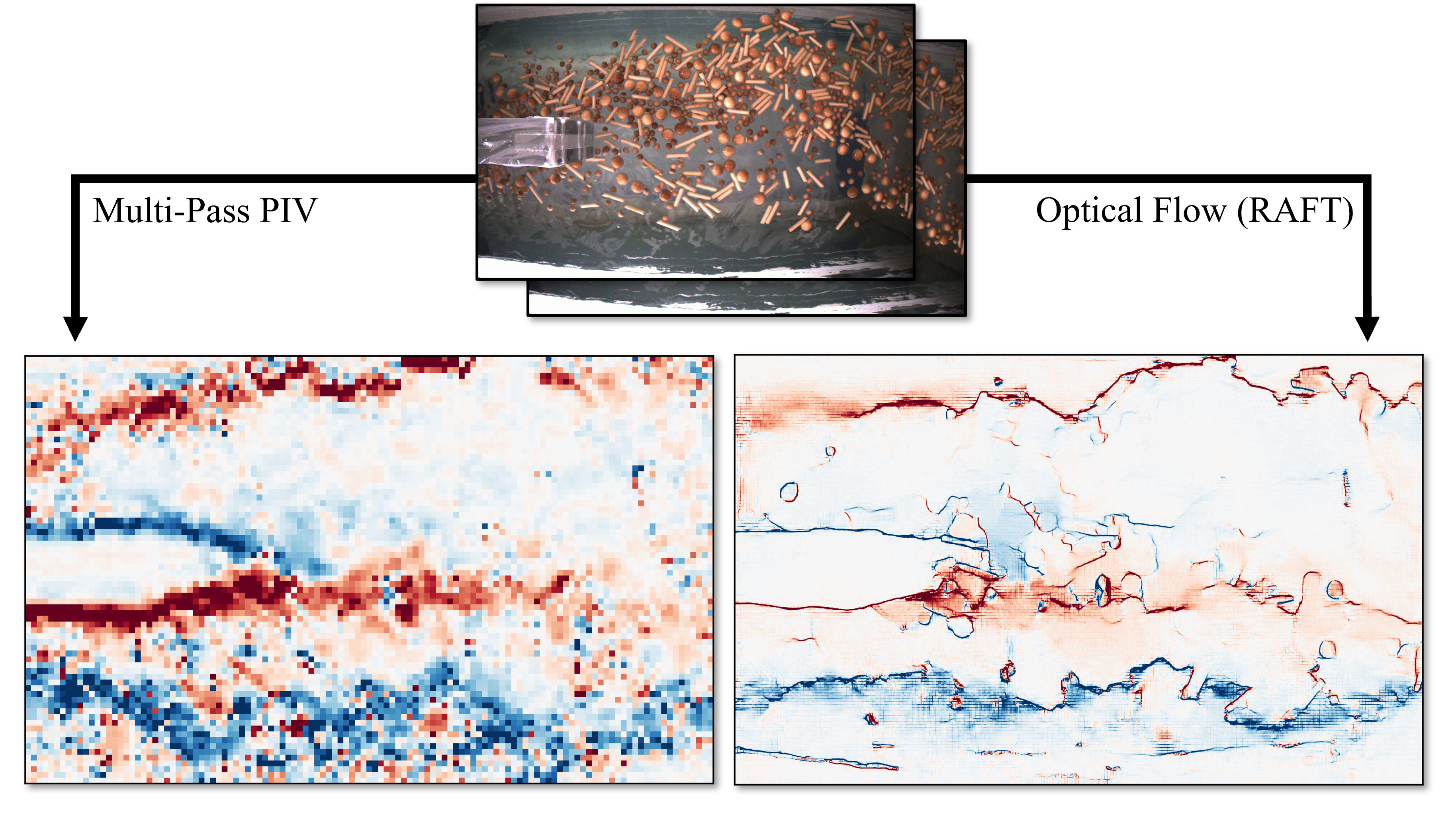}
    \caption{\edit{Computing flow gradients from an image sequence using standard procedures from experimental fluid mechanics (bottom left) and from computer vision (bottom right).  The algorithm used on the left is multi-pass PIV~\cite{Raffel.Kompenhans_PIVbook_2018}, and the algorithm used on the right is RAFT~\cite{TeedDeng_RAFT_2020}.  The results display vorticity ($\partial v/ \partial x - \partial u/ \partial y$).  From them it is clear that existing methods are not suited for spatial flow gradient estimation.  In both cases, there is significant noise and the existence of many spurious features.}}
    \label{fig:RAFT_vs_PIV}
\end{figure*}

\edit{
Flow gradients were estimated on the experimental images using conventional syntactic velocimetry approaches from fluid mechanics and computer vision.  The two algorithms considered are multi-pass particle image velocimetry (PIV)~\cite{Raffel.Kompenhans_PIVbook_2018} computed using the open-source package OpenPIV~\cite{Liberzon.Zimmer_OpenPIV_2021}, and  RAFT for optical flow~\cite{TeedDeng_RAFT_2020} using pre-trained weights available at the associated software repository.   The results are displayed in Figure \ref{fig:RAFT_vs_PIV}, where out-of-plane vorticity $\omega_z = \partial v/ \partial x - \partial u/ \partial y$ is displayed as an indicator of the gradient.  
}

\edit{
It is clear by the results displayed in Figure \ref{fig:RAFT_vs_PIV} that conventional motion estimation techniques are not well suited to approximate velocity gradients.  In both examples, false gradients are apparent at the top and bottom edges of the debris cluster (a red line near the top of the image and a blue line near the bottom).  These appear since PIV and optical flow are syntactic motion estimators, and are therefore agnostic to semantic information about the flow.  Rather, they attempt to identify motion in all parts of the image based on correlation or pixel intensity conservation.  Therefore, the algorithms perceive a strong false gradient between the cluster of debris (where motion is evident) to the surrounding flow without tracers (where motion exists, but is invisible to the camera).  There is also a significant amount of noise present in both analyses, which stems from the complexity of the images being analyzed and the presence of non-flow features such as reflections.  Finally, as a result of the poor quality of estimated velocity gradients, computing finite-time metrics like FTLE or LAVD using PIV or optical flow is challenging.  
}

\subsection{Implementation of the Proposed Method}
\edit{The methods discussed in the previous sections allow for flow gradients to be reliably estimated.  In this case study, the Mask-RCNN architecture \cite{He.Girschick_Mask-RCNN_2017} is employed for object detection and a rudimentary template matching scheme developed in-house is used for object tracking.  More details on the implementation of detection and tracking are provided below.}

\subsubsection{Stage 1: Detection}
\edit{
The Mask-RCNN model architecture was selected for the case study because it allows for full masks, and therefore accurate centroids, to be computed.  The model was implemented using PyTorch, where a pre-trained ResNet50 \cite{He.Sun_ResNet_2016} backbone was incorporated.  While three classes of particles existed in the flow, a custom head was built to identify all three as a single debris class.  Identifying a single class, however, is not necessary and may not be desirable in some cases.  The generality of large detection models enables studies to be performed on multiple classes of tracers at once.
}

Training data was collected using a custom app built around the Segment Anything Model (SAM) \cite{Kirilov.etal_SAM_2023}, which allowed for many precise masks to be identified quickly in training images.  In total, 165 training images were used with $>3000$ identified masks.  The custom model was trained for 100 epochs using a 120-to-45 training-to-testing split.  Stochastic gradient descent was used as the optimizer with initial learning rate of $0.0005$, momentum of $0.9$, and $L_2$ regularization of $0.0005$.  A scheduler reduced the learning rate by a factor of 10 every 10 epochs.  

Because the tracer particles were localized in crowds and were relatively small compared to the frame size, a windowing scheme was used to achieve improved detection results in each snapshot.  For the results discussed below, a $400\times 400$ pixel window with $10\%$ overlap was used for detections.  Any overlapping masks were consolidated into single particle masks, which were then used to identify the tracer centroids.  Figure \ref{fig:DetectionAndTracking}a shows an example of the windowing procedure applied to a sample image from the debris flow. 

\subsubsection{Stage 2: Tracking}

Object tracking was accomplished using a naive template matching scheme involving a forward search for the tracers identified in the first image of each image pair in the sequence.  A window around the identified tracer was specified as a template whose greatest correlation was found within a larger window in the subsequent frame.  If the peak of the correlation was found to be near a detection in the second frame, the detection was appended to the trajectory.  Velocity and acceleration constraints were employed to ensure that non-physical trajectories could not be created.  Once trajectories were computed, the camera calibration and homography were applied to the trajectory data.

\subsubsection{Stage 3: Structure Identification}
Gradient estimation and structure identification was performed using LGR according to the procedure outlined section \ref{sec:estimating dynamical features from trajectories}.  Equation \ref{eq:kernelweightedregression} was employed between each time step using the $k=15$ nearest neighbors of each tracer to populate the $\mb{X}_{t_0}$ and $\mb{X}_t$ matrices.  The kernel matrix $\mb{K}$ was populated on the diagonal according to
\begin{equation}
    \mb{K}_{ii} = e^{ -\frac{\Delta \mb{x}_i(t_0)^2}{2 s^2}},
    \label{eq:radialGaussian}
\end{equation}
with $s=0.03$ meters and set to zero otherwise.  A small regularization constant $\gamma$ was used to ensure numerical stability. 

\subsubsection{Visualization Approach}
\begin{figure*}[t!]
    \centering
    \includegraphics[width=1\textwidth]{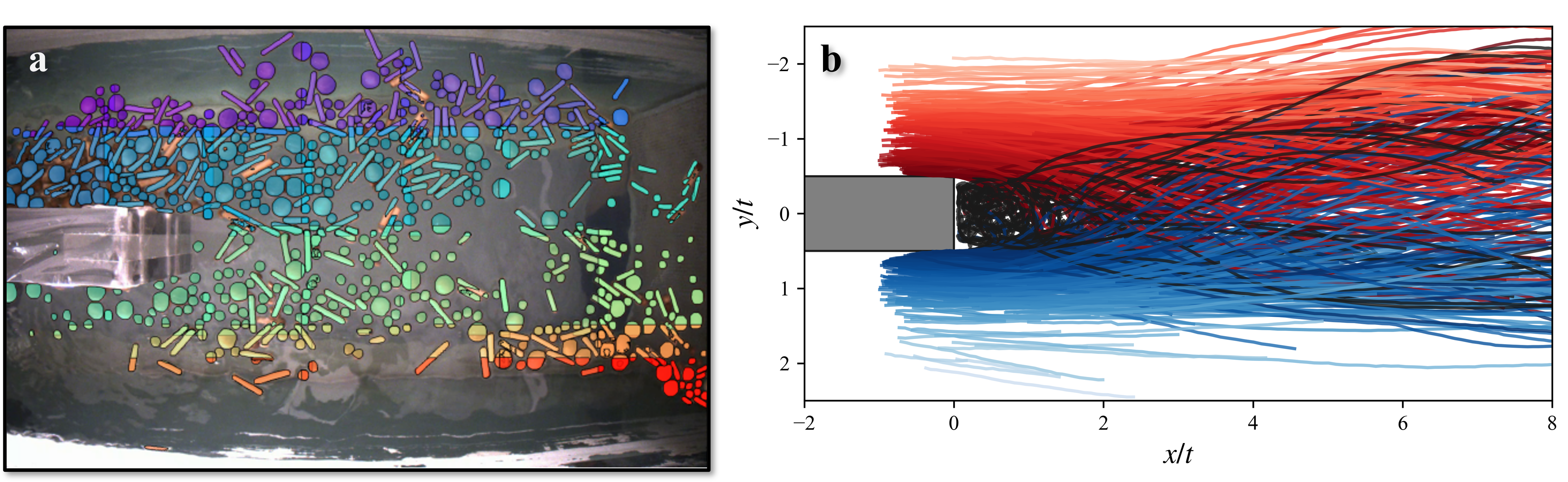}
    \caption{Sample results from the detection and tracking stages of the case study.  (a) A sample frame displaying detections using windowing with 10\% overlap.  Color indicates the windowed sub-image where detections were made.  Rods overlapping the boundaries of windows have not yet been corrected, but will be in further steps.  (b) A subset of trajectories length 50 or longer from the small birch spheres dataset.  All trajectories start in $x/t \in [-1, 1.25]$.  Color is determined by $y/t$.  If the first instance in a trajectory is in $y/t \in [-0.5, 0.5]$ it is colored black.}
    \label{fig:DetectionAndTracking}
\end{figure*} 
\label{ssec:visualization approach}
To visualize the data, each tracer centroid was provided as a query point to SAM \cite{Kirilov.etal_SAM_2023}, which identified its mask.  Each mask was overlaid on the original image with partial transparency and colored according to the value of the chosen metric.  This was done for vorticity, FTLE, and LAVD.  

\subsection{\edit{Results}}
Since the performance of LGR depends on the performance of identification and tracking, results from each stage of the process will be discussed in order.  Figure \ref{fig:DetectionAndTracking} presents sample results of detection and tracking, and Figure \ref{fig:LGR_Results} presents sample frames with various metrics computed.  Performance is considered for debris conditions with 1.\ only small spheres in the flow, and 2.\ all debris present.  Videos of the results are provided in the supplementary materials.

\subsubsection{Detection and Tracking}
The detection architecture discussed above was able to consistently identify a majority of the particles in each frame of the analyzed videos.  Unsurprisingly, the spheres were better detected than the rods were.  This seems to be due to the tendency of the rods to align lengthwise into groups, which were commonly undetected or misidentified.  Moreover, useful detections were not possible when the detector was applied to the full size image.  Only when the image was windowed were the tracers reliably segmented.  An example of a detected frame is shown in Figure \ref{fig:DetectionAndTracking}a.

Though the approach to object tracking is simple, it was able to construct a sufficient number of tracks for reliable LGR computation. Due to missed detections, many trajectories were split into segments, leading to heavy-tailed distributions of trajectory length.  This was pronounced for the data using all wooden debris, since detections were more frequently missed under those conditions.  Despite the truncated trajectories, sufficiently many tracks were recorded for successful application of LGR.  Figure \ref{fig:DetectionAndTracking}b shows a selection of the recorded tracks with 50 or more detections.

\subsubsection{Flow Feature Identification}
\begin{figure*}[t!]
    \centering
    \includegraphics[width=1\textwidth]{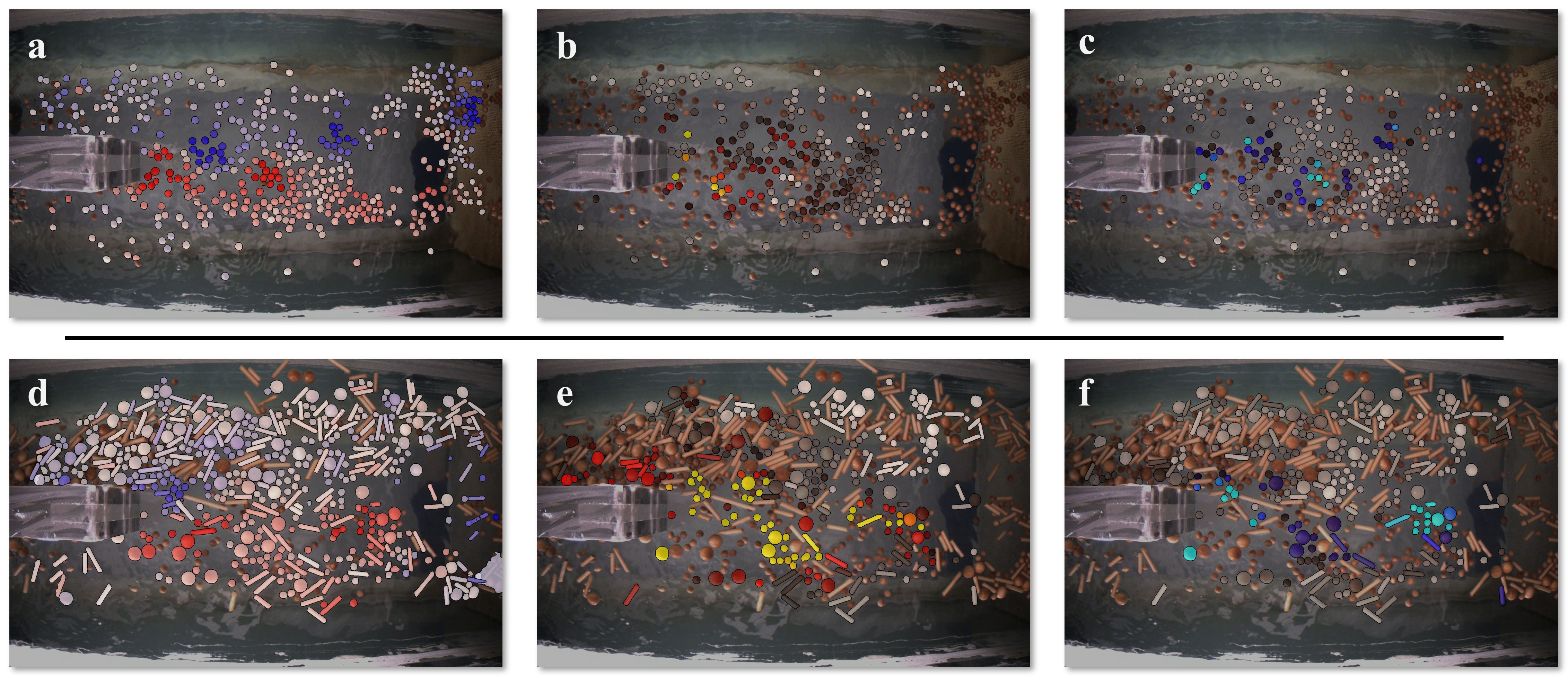}
    \caption{Results of the proposed analysis on the debris flow data.  Top: Small birch spheres as tracers.  Bottom:  All birch debris as tracers.  Vorticity is computed in (a) and (d), where red indicates positive value (counterclockwise rotation) and blue indicates negative value.  Finite-time stretching (FTLE) is presented in (b) and (e), where yellow indicates large value and white is zero, and finite-time rotation (LAVD) is given in (c) and (f), where bright blue indicates large values.}
    \label{fig:LGR_Results}
\end{figure*}

The performance of LGR is first demonstrated through the computation of vorticity, or the instantaneous rate of rotation.  Figure \ref{fig:LGR_Results}a presents the vorticity for data containing only small birch spheres as tracers.  Here, the vortex shedding that occurs at the trailing edge of the plate is clearly visible in the computed vorticity values.  In the given frame, there are three clockwise vortices (blue clusters) which alternate with counterclockwise vortices (red clusters).  The vorticity in the data set using all debris is presented in Figure \ref{fig:LGR_Results}d, where the inclusion of large spheres and rods increases the density of tracers on the water surface and thereby impedes some tracer motion through particle interactions.  This is clearly seen in the vorticity data, where there are fewer vortices (clusters of either red or blue) visible.  

Finite time metrics can also be used to identify patterns in the flow of detected tracers.  Measurements of maximal stretching (FTLE) are presented in Figures \ref{fig:LGR_Results}b and \ref{fig:LGR_Results}e for the spherical tracers and for all tracer classes respectively.  A large value of FTLE (red to yellow masks on the image) indicates that surrounding tracers will separate over the duration for which the computation is performed.  The theory of LCS suggests that ridges in the FTLE field are codimension-1 material surfaces that act as barriers of flow transport \cite{Shadden.Marsden_FTLEProperties_2005}.  Because they are infinitesimally thin, FTLE ridges are difficult to identify from sparse data such as that in Figure \ref{fig:LGR_Results}~\cite{Harms.McKeon_LGR_2023}  Nevertheless, identifying regions of trajectory divergence remains useful even when sharp ridges may be difficult to discern.  

LGR also enables measurements of cumulative rotation through computation of LAVD.  The LAVD results for the case study are displayed in figures \ref{fig:LGR_Results}c and \ref{fig:LGR_Results}f for the spherical tracers and for all tracer classes respectively.  If a tracer exhibits a large value of LAVD (bright blue masks), then surrounding tracers will rotate around the examined trajectory.  Unlike FTLE fields, LAVD reveals volumetric regions of the material that experience significant rotation. 
Therefore, clusters of tracers with large LAVD values can be instructive for identifying coherent structures in the dynamical system.  Such clusters are clearly visible in both figures \ref{fig:LGR_Results}c and \ref{fig:LGR_Results}f.

\section{\edit{Field Test: Pond Debris}}
\label{sec:field test: pond debris}
\edit{
To extend this approach to flows outside of a controlled environment, a field experiment was conducted at the turtle ponds adjacent to the Graduate Aerospace Laboratories (GALCIT) at Caltech.  The turtle ponds are a landscaping feature on Caltech's campus where leaves, bubbles, and other debris are often seen on the pond surface.  Small streams and waterfalls drive the motion of the fluid, which is visible in the trajectories of the floating debris.  In this experiment, these are filmed and used to compute flow quantities.  No artificial seeding is used.
}
\subsection{\edit{Experimental Setup}}
\edit{
To emphasize the robustness of the proposed method, the field experiment on the turtle ponds was conducted using common hobbyist recording equipment.  All images were recorded using a Nikon D800 DSLR camera commonly used for digital photography.  When set to record videos, the D800 can sample frames with $1920 \times 1080$ pixel resolution at 30 frames per second.  Videos were recorded at multiple locations around the turtle ponds, and on different days.  Two lenses were used during filming: a Nikon AF Nikkor 35mm f/2D lens and a Nikon AF Nikkor 50mm f/1.8D lens.  The results discussed in this section were taken using the 35mm lens on a day when the natural seeding in the ponds was particularly heavy.  
}

\begin{figure*}[t!]
    \centering
    \includegraphics[width=0.8\textwidth]{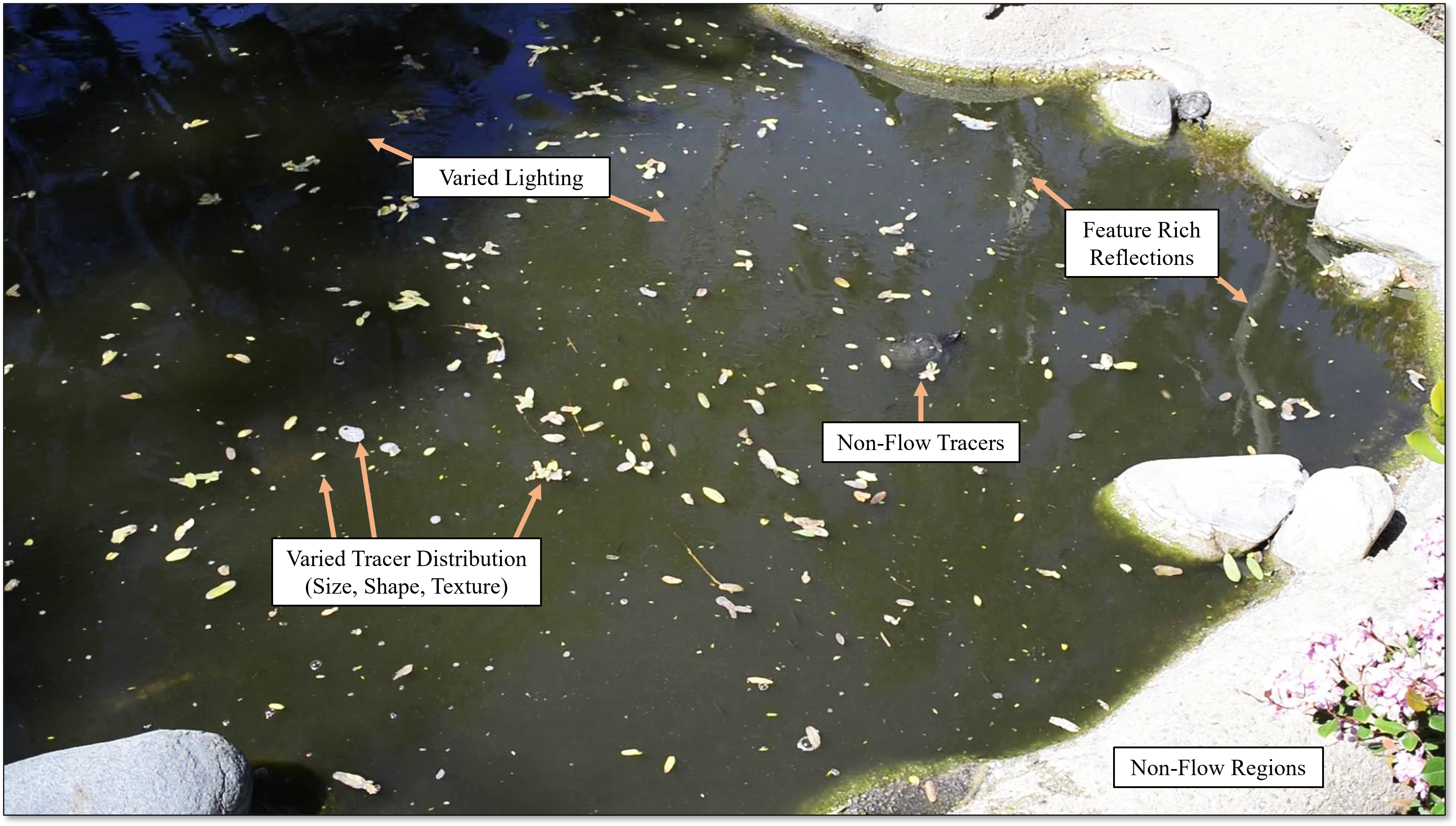}
    \caption{\edit{An example image from the turtle ponds data set being analyzed in section \ref{sec:field test: pond debris}.  Image features which complicate normal motion and gradient estimates are highlighted.  }}
    \label{fig:TP_image}
\end{figure*}
\edit{
An example image from the turtle pond data set is annotated in Figure \ref{fig:TP_image} to emphasize the feature-rich complexity of the analyzed images.  The scene being considered includes shadows and variations in lighting, multiple forms of legitimate tracers (bubbles, leaves, and other debris), illegitimate tracers (including turtles and fish),  feature rich reflections from the surrounding flora which ripple in the motion of the water, and regions of the image which are not part of the flow.  All of these attributes severely complicate the estimation of velocities and, especially, of spatial flow gradients.  
}

\edit{
Calibration was performed in the same manner as with the laboratory case study, using the same ChArUco board as in the previous example.  First, the camera matrix was calibrated using a sequence of images containing the board.  Next, images were recorded of the calibration board in the plane of the flow (on the pond surface).  Since the camera could not be situated directly above the flow as in the lab study, the in-plane images served not only to appropriately scale the flow, but also to orient the camera to an overhead position.  This was accomplished by fitting a homography matrix to one of the in-plane images. 
}

\subsection{\edit{Implementation}}
\edit{
The same algorithms were used on the turtle pond data as were used in the laboratory experiment.   For detection, the Mask-RCNN~\cite{He.Girschick_Mask-RCNN_2017} detector architecture was employed to segment legitimate tracers and find their centroids.  110 training images with all contained tracer masks were collected using the same custom app built on SAM~\cite{Kirilov.etal_SAM_2023}, and the detection model was trained for 150 epochs using the same optimization parameters as in the previous case.  Windowing was again used to improve the quality of detections.  In the turtle pond images, detections were made inside $400\times 400$ pixel tiles with $25\%$ overlap.  
}

\edit{
Trajectories were tracked using the same simple correlation-based approach described in the previous section with some slight modifications.  First, to reduce the number stored trajectories and to improve the average quality of tracks, only trajectories with length 5 or greater were kept.  Additionally, the pond data and the detection scheme used produced trajectories that were slightly noisy.  To mitigate this, signal filtering was implemented on the identified trajectories.  A median filter with a 5 sample kernel length was used to remove outliers, followed by a Gaussian smoothing filter with a 10 sample kernel length.  These filters both padded the trajectory ends with duplicates.  
}
\subsection{\edit{Results}}
\edit{
Flow gradients were estimated using LGR with radial Gaussian kernel weighting.  For each tracer at each time step, the 25 nearest neighbors that persisted from the current frame to the next were used to regress the deformation operator by equation \ref{eq:kernelweightedregression}.  The weighting was defined using equation \ref{eq:radialGaussian}, where the standard deviation was set to $s=0.6$ meters.  Velocity gradients were then estimated using equation \ref{eq:gradv_approx}.  Finite-time analyses were computed over an 8 second interval using the estimated gradients on all particles that persist throughout the entire interval.   The full length of the recording is nearly 64 seconds.  
}
\begin{figure*}[t!]
    \centering
    \includegraphics[width=1\textwidth]{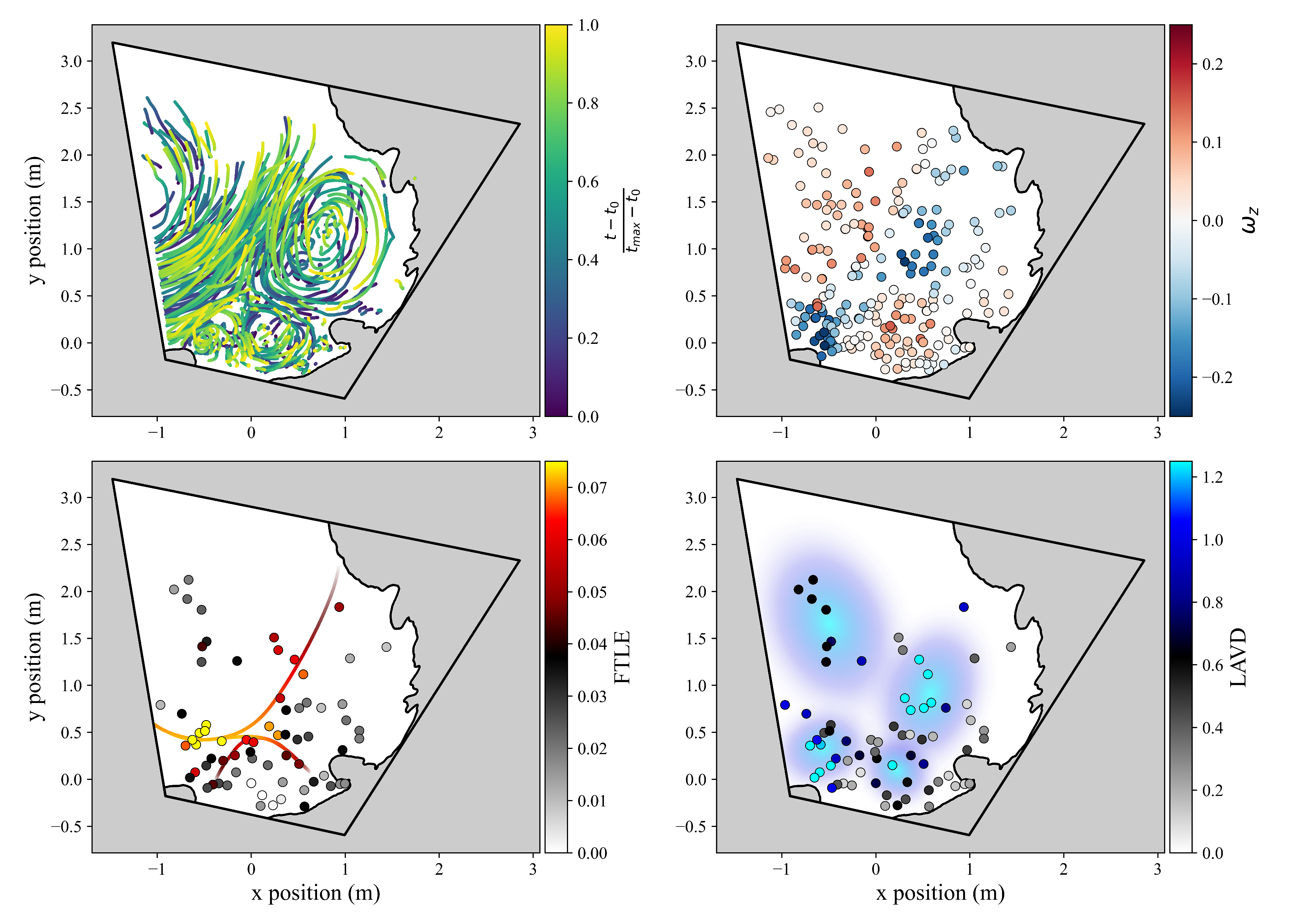}
    \caption{\edit{Results from the field data collected in the Caltech turtle ponds.  (a) Sample of identified trajectories length 200 ($6$ and $2/3$ seconds) or greater calibrated to world coordinates and colored by nondimensional time $t^*$ within the recording. (b) Vorticity computed at $t^*=0.29$ and displayed in world coordinates.  (c) Forward-time FTLE computed over 8 seconds from $t^*=0.29$ to $t^*=0.41$ and plotted in world coordinates.  The curve indicating the hyperbolic structure is artificially added in a graphical editor for emphasis.  (d) LAVD computed over 8 seconds from $t^*=0.29$ to $t^*=0.41$ and plotted in world coordinates.  The blue shaded regions indicating Lagrangian vortices are artificially added in a graphical editor for emphasis.}}
    \label{fig:TP_results}
\end{figure*}

\edit{
Example results from the field experiment in the Caltech turtle ponds are provided in Figure \ref{fig:TP_results}.  Figure \ref{fig:TP_results}a displays the subset of identified trajectories with length of $200$ or more snapshots ($6$ and $2/3$ seconds or longer).  The coloring scheme indicates the relative position in time of the trajectory instance within the recording using a nondimensional time defined as $t^* = \frac{t-t_0}{t_{max}-t_0}$, where $t_0$ is the time at the first frame and $t_{max}$ is the time at the last frame.  Figure \ref{fig:TP_results}b displays the instantaneous vorticity at $t^* = 0.29$ for all observed tracers which were tracked over one time step, \ref{fig:TP_results}c displays the forward-time FTLE computed on all trajectories persisting for 8 seconds on the interval $t^* = [0.29,\, 0.41]$, and \ref{fig:TP_results}d displays the LAVD computed on the same trajectories as in \ref{fig:TP_results}c.  Fewer measurements are available in Figures \ref{fig:TP_results}c and \ref{fig:TP_results}d since many of the trajectories were cut short at some point in the time interval $t^* = [0.29,\, 0.41]$.  All data is presented in the planar, dimensional coordinate system of the flow.  The trapezoidal boundary which contains the data represents the full domain of the image sensor in world coordinates.   
}

\edit{
From the displayed results, flow features can be identified and quantified.  Judging from the visualization of the tracer trajectories, there are estimated four vortices present in the flow.  From the plot of vorticity in Figure \ref{fig:TP_results}b, it is evident that there are two clockwise vortices (shown in blue) and two counter-clockwise ones (shown in red).  These vortices are further expressed in the finite-time context through the LAVD in Figure \ref{fig:TP_results}d, where the displayed color values are presented in radians rotated over the 8 second interval.  Shading has been added to emphasize the position of these vortices.  By this analysis, it is apparent that the clockwise vortices are the largest and strongest observed in the flow at this time snapshot.  The vorticity and LAVD results are complemented by those of the FTLE in Figure \ref{fig:TP_results}c where domains of maximal trajectory separation are identified.  Ridges in these fields should occur at the interface between counter-rotating vortices, which is largely observed by the data.  Sketched separatrices have been added in a graphical editor to highlight where these ridges are indicated by the data.  
}

\edit{
Considering the various metrics computed on the flow together proves to be more enlightening than considering any one by itself.  For example, observing the trajectories yields qualitative information about the dynamics of the flow, but vorticity quantifies it at an instantaneous snapshot.  LAVD and FTLE extend the understanding of the flow to finite intervals.  Thus, one of the advantages of using LGR in studies such as this is that enables many gradient-based metrics simultaneously.  
}

\section{Discussion}
\label{sec:discussion}
\edit{
The two examples presented above illustrate various attributes of the proposed approach, which are now discussed.  
}

\subsection{\edit{Reliable Gradients from Difficult Data}}
\edit{
The methods introduced in this work have been designed with the intent of enabling the computation of flow gradients and related metrics in dynamical systems which are typically intractable to gradient-based analyses.  Both the experimental case study of section \ref{sec:case study: experimental debris flow} and the field study in section \ref{sec:field test: pond debris} represent data sets where flow gradient estimation is extremely difficult when using existing approaches.  By implementing the methods of sections \ref{sec:estimating dynamical features from trajectories} and \ref{sec:detection and tracking}, reliable gradients and associated metrics could directly computed with little additional processing. 
}

\subsection{Limitations}
Missed detections resulting in truncated trajectories represent the primary barrier to improved LGR performance.  This is especially true when computing finite-time metrics like the FTLE and LAVD, which require the existence of primary trajectories for the entire computation interval.  In the laboratory case study, while there were missed detections for both the small spheres and all debris cases, they were more frequent when using all of the debris.  Rods, and especially groups of aligned rods, were less likely to be detected than the spheres.  This suggests that improvements can be made by training on more data.  Additionally, as the algorithm is currently implemented, the detections are independent in each frame.  However, significant improvement could be made to the LGR outcome if tracking techniques that incorporate detections from previous frames such as SORT \cite{Bewley.Upcroft_SORT_2016} or deepSORT \cite{Wojke.Paulus_DeepSORT_2017} were implemented for trajectory construction.  Thus, it seems that there are many algorithmic improvements that can be quickly implemented to improve performance.  

Moreover, while real-time implementation of the LGR pipeline is a long-term goal of this work, it is not yet achievable.  With the current implementation and modest hardware, processing one of the debris flow videos requires on the order of hours of computation.  This time could be greatly reduced, however, if one were to optimize the code for better performance on a GPU.  Moreover, if faster detection and tracking algorithms are implemented, more speedup will be achieved.  Still, this type of analysis is only viable for offline studies in most cases at the moment.  

\subsection{\edit{Translation and Rotation Invariant Analyses}}
\edit{
While not all gradient-based metrics are invariant to change-of-frame (vorticity, for example, is sensitive to observer rotation), the ability to reliably compute gradients enables objective metrics that are insensitive to observer dynamics.  Metrics such as these are the subject of the study of Lagrangian coherent structures (LCS), which utilizes gradients estimated from tracer trajectories to achieve objective analyses.  FTLE and LAVD are examples of such objective analyses, and will yield identical results along trajectories regardless of the observer motion.  
}

\edit{
LCS analyses are often limited by the availability of observable tracers.  The methods proposed in this work reduce these limitations by identifying trajectories in complex, feature-rich images using deep object detection and tracking algorithms and by estimating flow gradients from sparse gradients with LGR.  These two advances allow objective gradient-based analyses to be extended to broader classes of dynamical systems.  This type of analysis may prove useful in the modern autonomy context where vehicles and drones must safely respond to the dynamical features of their environment as they navigate.  
}

\subsection{\edit{Opportunity for Multi-Class Analysis}}
\edit{
Though it was left unexplored in this work, the proposed method enables the dynamics of multiple classes of objects to be simultaneously studied.  For example, if this approach were pursued on the laboratory case study of section \ref{sec:case study: experimental debris flow}, the dynamics of spheres might be compared to that of the rods through a single analysis.  This principal could be extended to studies involving bubbles and sediment, vehicles and pedestrians, predators and prey, or bacteria and blood cells, among many other combinations.  
}

\subsection{\edit{Modular Framework for Continuous Improvement}}
\edit{
The framework developed in this work has been modularly constructed to accommodate the rapid advancement of object detection and tracking capabilities.  The detection and tracking schemes used in the examples above are relatively naive given the current state of the field, and therefore serve as an apt demonstration of the framework as a whole; if gradients can be successfully computed using simple detection and tracking schemes, then more advanced methods will be able to improve upon the results.   
}

\subsection{\edit{Lenient Hardware Requirements}}
\edit{
The methods introduced in this work can be easily implemented using affordable, readily available hardware.  All of the data presented in the experiments was collected using measurement systems costing less than $1000$ USD.  The cost of laboratory experiments, by contrast, can easily exceed orders of magnitude greater than this.  Moreover, due to the availability of large pre-trained computer vision models, detector training costs are modest and do not require more computational power than a desktop computer.  Therefore, the methods developed here are relatively democratic in the sense that they enable measurements to be made by hobbyists and others without access to specialized equipment.  
}

\section{Conclusions}
\label{sec:conclusions}

In this work, a modular tool for detecting dynamical features in systems of arbitrary tracked objects was developed and demonstrated.  Object trajectories are computed through deep detection and tracking algorithms and are supplied as inputs to Lagrangian gradient regression, which computes quantities of interest regarding the vector field that the dynamical objects evolve on.  The tool is built in a modular fashion, which enables new developments in object detection and tracking to be implemented as they become available.  The proposed method was tested on two cases meant to highlight the method's ability: an experimental surface tracer flow and a flow of debris on the surface of a pond.  These case studies contained flows in which gradients are quite challenging to measure and served to judge the merits of the described method.

\edit{
The proposed approach enables gradient-based analyses of motion and dynamics from affordably captured image sequences containing the collective movement of complex, semantic objects.  Standard approaches to motion estimation such as optical flow are not well-suited for gradient estimation in such data.  Features such as reflections, variation in lighting, sparse seeding of tracers, and others prevent their application.  Through the presented case studies, however, the proposed approach was seen to be effective at estimating gradients from feature-rich flow images.  Given the gradients, objective metrics of dynamical behavior insensitive to camera motion are readily computed.  Vorticity deviation, FTLE, and LAVD are examples of such objective metrics that have been widely successful in the fluid dynamics community and could find broad utility in other dynamical systems and computer vision applications. Another advantage of the proposed tool is that it allows for the motion of systems of distinct classes to be analyzed within the same image sequence (for example, vehicles and pedestrians, predators and prey, or red and white blood cells).
}

This approach is broadly generalizable to many applications.  As demonstrated, it is useful in the study of fluid dynamics from field data, but the analysis can be extended to other examples of object motion.  Any system which can be observed as a crowd of dynamic objects is legitimate.  Thus, potential candidates include mircofluidics, traffic and population flows, environmental flows, and herds and swarms, among others.  

\bibliographystyle{ieeetr}
\bibliography{Harms_et_al__2024__ArXiv}

\begin{thebibliography}{10}

\bibitem{Raffel.Kompenhans_PIVbook_2018}
M.~Raffel, C.~E. Willert, F.~Scarano, C.~J. K{\"a}hler, S.~T. Wereley, and
  J.~Kompenhans, {\em Particle Image Velocimetry: A Practical Guide}.
\newblock Cham: Springer International Publishing, 2018.

\bibitem{Haller_LCS_2015}
G.~Haller, ``{L}agrangian coherent structures,'' {\em Annual Review of Fluid
  Mechanics}, vol.~47, pp.~137--162, 2015.

\bibitem{Fortun.Kervrann_OFSurvey_2015}
D.~Fortun, P.~Bouthemy, and C.~Kervrann, ``Optical flow modeling and
  computation: A survey,'' {\em Computer Vision and Image Understanding},
  vol.~134, pp.~1--21, 2015.

\bibitem{Zhai.Kong_OFSurvey_2021}
M.~Zhai, X.~Xiang, N.~Lv, and X.~Kong, ``Optical flow and scene flow
  estimation: A survey,'' {\em Pattern Recognition}, vol.~114, p.~107861, 2021.

\bibitem{BakerMatthews_L-KSurvey_2004}
S.~Baker and I.~Matthews, ``Lucas-{K}anade 20 years on: A unifying framework,''
  {\em International journal of computer vision}, vol.~56, pp.~221--255, 2004.

\bibitem{HornSchunk_HSOF_1981}
B.~K.~P. Horn and B.~G. Schunck, ``Determining optical flow,'' {\em Artificial
  intelligence}, vol.~17, no.~1-3, pp.~185--203, 1981.

\bibitem{Dosovitskiy.Brox_FlowNet_2015}
A.~Dosovitskiy, P.~Fischer, E.~Ilg, P.~Hausser, C.~Hazirbas, V.~Golkov, P.~Van
  Der~Smagt, D.~Cremers, and T.~Brox, ``Flownet: Learning optical flow with
  convolutional networks,'' in {\em Proceedings of the IEEE international
  conference on computer vision}, pp.~2758--2766, 2015.

\bibitem{Ilg.Brox_FlowNet2_2017}
E.~Ilg, N.~Mayer, T.~Saikia, M.~Keuper, A.~Dosovitskiy, and T.~Brox, ``Flownet
  2.0: Evolution of optical flow estimation with deep networks,'' in {\em
  Proceedings of the IEEE conference on computer vision and pattern
  recognition}, pp.~2462--2470, 2017.

\bibitem{Sun.Kautz_PWC-Net_2018}
D.~Sun, X.~Yang, M.-Y. Liu, and J.~Kautz, ``{PWC}-{N}et: {CNN}s for optical
  flow using pyramid, warping, and cost volume,'' in {\em Proceedings of the
  IEEE conference on computer vision and pattern recognition}, pp.~8934--8943,
  2018.

\bibitem{TeedDeng_RAFT_2020}
Z.~Teed and J.~Deng, ``{RAFT}: Recurrent all-pairs field transforms for optical
  flow,'' in {\em Computer Vision--ECCV 2020: 16th European Conference,
  Glasgow, UK, August 23--28, 2020, Proceedings, Part II 16}, pp.~402--419,
  Springer, 2020.

\bibitem{Xu.Li_DeepMOTSurvey_2019}
Y.~Xu, X.~Zhou, S.~Chen, and F.~Li, ``Deep learning for multiple object
  tracking: a survey,'' {\em IET Computer Vision}, vol.~13, no.~4,
  pp.~355--368, 2019.

\bibitem{Luo.Kim_MOTReview_2021}
W.~Luo, J.~Xing, A.~Milan, X.~Zhang, W.~Liu, and T.-K. Kim, ``Multiple object
  tracking: A literature review,'' {\em Artificial intelligence}, vol.~293,
  p.~103448, 2021.

\bibitem{Voigtlaender.Leibe_MOTS_2019}
P.~Voigtlaender, M.~Krause, A.~Osep, J.~Luiten, B.~B.~G. Sekar, A.~Geiger, and
  B.~Leibe, ``{MOTS}: {M}ulti-object tracking and segmentation,'' in {\em
  Proceedings of the ieee/cvf conference on computer vision and pattern
  recognition}, pp.~7942--7951, 2019.

\bibitem{Su.Luo_TransTrack_2020}
P.~Sun, J.~Cao, Y.~Jiang, R.~Zhang, E.~Xie, Z.~Yuan, C.~Wang, and P.~Luo,
  ``Transtrack: {M}ultiple object tracking with transformer,'' {\em arXiv
  preprint arXiv:2012.15460}, 2020.

\bibitem{Meinhardt.Feichtenhofer_Trackformer_2022}
T.~Meinhardt, A.~Kirillov, L.~Leal-Taixe, and C.~Feichtenhofer, ``Trackformer:
  {M}ulti-object tracking with transformers,'' in {\em Proceedings of the
  IEEE/CVF conference on computer vision and pattern recognition},
  pp.~8844--8854, 2022.

\bibitem{Liu.Yu_GraphGSM_2020}
Q.~Liu, Q.~Chu, B.~Liu, and N.~Yu, ``{GSM}: {G}raph {S}imilarity {M}odel for
  {M}ulti-{O}bject {T}racking.,'' in {\em IJCAI}, pp.~530--536, 2020.

\bibitem{Quach.Luu_DyGlip_2021}
K.~G. Quach, P.~Nguyen, H.~Le, T.-D. Truong, C.~N. Duong, M.-T. Tran, and
  K.~Luu, ``Dyglip: A dynamic graph model with link prediction for accurate
  multi-camera multiple object tracking,'' in {\em Proceedings of the IEEE/CVF
  Conference on Computer Vision and Pattern Recognition}, pp.~13784--13793,
  2021.

\bibitem{Guo.Shen_GraphAttention_2021}
D.~Guo, Y.~Shao, Y.~Cui, Z.~Wang, L.~Zhang, and C.~Shen, ``Graph attention
  tracking,'' in {\em Proceedings of the IEEE/CVF conference on computer vision
  and pattern recognition}, pp.~9543--9552, 2021.

\bibitem{Chu.Yu_OnlineMOTAttention_2017}
Q.~Chu, W.~Ouyang, H.~Li, X.~Wang, B.~Liu, and N.~Yu, ``Online multi-object
  tracking using {CNN}-based single object tracker with spatial-temporal
  attention mechanism,'' in {\em Proceedings of the IEEE international
  conference on computer vision}, pp.~4836--4845, 2017.

\bibitem{Zhan.Xu_CrowdSurvey_2008}
B.~Zhan, D.~N. Monekosso, P.~Remagnino, S.~A. Velastin, and L.-Q. Xu, ``Crowd
  analysis: a survey,'' {\em Machine Vision and Applications}, vol.~19,
  pp.~345--357, 2008.

\bibitem{Sanchez.Herrera_DeepCrowdAnalysis_2020}
F.~L. S{\'a}nchez, I.~Hupont, S.~Tabik, and F.~Herrera, ``Revisiting crowd
  behaviour analysis through deep learning: Taxonomy, anomaly detection, crowd
  emotions, datasets, opportunities and prospects,'' {\em Information Fusion},
  vol.~64, pp.~318--335, 2020.

\bibitem{cheriyadatRadke_DominantMotionInCrowds_2008}
A.~M. Cheriyadat and R.~J. Radke, ``Detecting dominant motions in dense
  crowds,'' {\em IEEE Journal of Selected Topics in Signal Processing}, vol.~2,
  no.~4, pp.~568--581, 2008.

\bibitem{Verma.Kulkarni_COVIDCrowdDetection_2021}
N.~Verma, S.~Patil, B.~Sinha, and V.~Kulkarni, ``Object {D}etection for {COVID}
  {R}ules {R}esponse and {C}rowd {A}nalysis,'' in {\em 2021 Innovations in
  Power and Advanced Computing Technologies (i-PACT)}, pp.~1--6, IEEE, 2021.

\bibitem{Pouw.Corbetta_CrowdManPhysicalDistancing_2020}
C.~A. Pouw, F.~Toschi, F.~van Schadewijk, and A.~Corbetta, ``Monitoring
  physical distancing for crowd management: Real-time trajectory and group
  analysis,'' {\em PloS one}, vol.~15, no.~10, p.~e0240963, 2020.

\bibitem{Schanz.Schroder_ShakeTheBox_2016}
D.~Schanz, S.~Gesemann, and A.~Schr{\"o}der, ``Shake-the-box: {L}agrangian
  particle tracking at high particle image densities,'' {\em Experiments in
  Fluids}, vol.~57, p.~70, May 2016.

\bibitem{SchroderSchanz_PTVReview_2023}
A.~Schr{\"o}der and D.~Schanz, ``{3D} {L}agrangian particle tracking in fluid
  mechanics,'' {\em Annual Review of Fluid Mechanics}, vol.~55, pp.~511--540,
  2023.

\bibitem{AllshousePeacock_LagrangianBasedMethods_2015}
M.~R. Allshouse and T.~Peacock, ``{L}agrangian based methods for coherent
  structure detection,'' {\em Chaos: An Interdisciplinary Journal of Nonlinear
  Science}, vol.~25, p.~097617, Sept. 2015.

\bibitem{Haller_LCStextbook_2023}
G.~Haller, {\em Transport Barriers and Coherent Structures in Flow Data:
  Advective, Diffusive, Stochastic and Active Methods}.
\newblock Cambridge University Press, 2023.

\bibitem{Gurtin.Anand_ContinuumMechanics_2010}
M.~E. Gurtin, E.~Fried, and L.~Anand, {\em The Mechanics and Thermodynamics of
  Continua}.
\newblock 32 Avenue of the Americas, New York NY 10013-2473, USA: {Cambridge
  University Press}, 2010.

\bibitem{OlascoagaHaller_ForecastingPollutionPatterns_2012}
M.~J. Olascoaga and G.~Haller, ``Forecasting sudden changes in environmental
  pollution patterns,'' {\em Proceedings of the National Academy of Sciences},
  vol.~109, pp.~4738--4743, Mar. 2012.

\bibitem{Nolan.Ross_app.bio_2020}
P.~J. Nolan, H.~Foroutan, and S.~D. Ross, ``Pollution transport patterns
  obtained through generalized {Lagrangian} coherent structures,'' {\em
  Atmosphere}, vol.~11, no.~2, p.~168, 2020.

\bibitem{PengDabiri_app.bio_2009}
J.~Peng and J.~O. Dabiri, ``Transport of inertial particles by {Lagrangian}
  coherent structures: application to predator--prey interaction in jellyfish
  feeding,'' {\em Journal of Fluid Mechanics}, vol.~623, pp.~75--84, 2009.

\bibitem{Shadden.Gerbeau_app.bio_2010}
S.~C. Shadden, M.~Astorino, and J.-F. Gerbeau, ``Computational analysis of an
  aortic valve jet with {Lagrangian} coherent structures,'' {\em Chaos: An
  Interdisciplinary Journal of Nonlinear Science}, vol.~20, no.~1, 2010.

\bibitem{Ahmed.Hanifatu_app.aero_2023}
D.~Ahmed, A.~Javed, M.~S. Uz~Zaman, M.~Mahsud, M.-N. Hanifatu, {\em et~al.},
  ``Efficient sensor location for {HVAC} systems using {Lagrangian} coherent
  structures,'' {\em Mathematical Problems in Engineering}, vol.~2023, 2023.

\bibitem{Haller.Encinas-Bartos_Quasi-ObjectiveDiagnostics_2021}
G.~Haller, N.~Aksamit, and A.~P. {Encinas-Bartos}, ``Quasi-objective coherent
  structure diagnostics from single trajectories,'' {\em Chaos: An
  Interdisciplinary Journal of Nonlinear Science}, vol.~31, p.~043131, Apr.
  2021.

\bibitem{Encinos-Bartos.Haller_objectiveEddyViz_2022}
A.~P. Encinas-Bartos, N.~O. Aksamit, and G.~Haller, ``Quasi-objective eddy
  visualization from sparse drifter data,'' {\em Chaos: An Interdisciplinary
  Journal of Nonlinear Science}, vol.~32, no.~11, 2022.

\bibitem{Harms.McKeon_LGR_2023}
T.~D. Harms, S.~L. Brunton, and B.~J. McKeon, ``Lagrangian gradient regression
  for the detection of coherent structures from sparse trajectory data,'' 2023.

\bibitem{Girschick_Fast-RCNN_2015}
R.~Girshick, ``Fast {R-CNN},'' in {\em Proceedings of the IEEE international
  conference on computer vision}, pp.~1440--1448, 2015.

\bibitem{He.Girschick_Mask-RCNN_2017}
K.~He, G.~Gkioxari, P.~Doll{\'a}r, and R.~Girshick, ``Mask {R-CNN},'' in {\em
  Proceedings of the IEEE international conference on computer vision},
  pp.~2961--2969, 2017.

\bibitem{Redmon.Farhadi_YOLO_2016}
J.~Redmon, S.~Divvala, R.~Girshick, and A.~Farhadi, ``You only look once:
  Unified, real-time object detection,'' in {\em Proceedings of the IEEE
  conference on computer vision and pattern recognition}, pp.~779--788, 2016.

\bibitem{Jiang.Ma_YOLOReview_2022}
P.~Jiang, D.~Ergu, F.~Liu, Y.~Cai, and B.~Ma, ``A review of yolo algorithm
  developments,'' {\em Procedia Computer Science}, vol.~199, pp.~1066--1073,
  2022.

\bibitem{Carion.Zagoruyko_DETR_2020}
N.~Carion, F.~Massa, G.~Synnaeve, N.~Usunier, A.~Kirillov, and S.~Zagoruyko,
  ``End-to-end object detection with transformers,'' in {\em European
  conference on computer vision}, pp.~213--229, Springer, 2020.

\bibitem{Zhuang.He_TransferLearning_2020}
F.~Zhuang, Z.~Qi, K.~Duan, D.~Xi, Y.~Zhu, H.~Zhu, H.~Xiong, and Q.~He, ``A
  comprehensive survey on transfer learning,'' {\em Proceedings of the IEEE},
  vol.~109, no.~1, pp.~43--76, 2020.

\bibitem{Bewley.Upcroft_SORT_2016}
A.~Bewley, Z.~Ge, L.~Ott, F.~Ramos, and B.~Upcroft, ``Simple online and
  realtime tracking,'' in {\em 2016 IEEE international conference on image
  processing (ICIP)}, pp.~3464--3468, IEEE, 2016.

\bibitem{Wojke.Paulus_DeepSORT_2017}
N.~Wojke, A.~Bewley, and D.~Paulus, ``Simple online and realtime tracking with
  a deep association metric,'' in {\em 2017 IEEE international conference on
  image processing (ICIP)}, pp.~3645--3649, IEEE, 2017.

\bibitem{GuckenheimerHolmes_NonlinearDynamicalBifurcation_1983}
J.~Guckenheimer and P.~Holmes, {\em Nonlinear Oscillations, Dynamical Systems,
  and Bifurcations of Vector Fields}, vol.~42 of {\em Applied Mathematical
  Sciences}.
\newblock {New York, NY}: Springer New York, 1983.

\bibitem{Shadden.Marsden_FTLEProperties_2005}
S.~C. Shadden, F.~Lekien, and J.~E. Marsden, ``Definition and properties of
  {L}agrangian coherent structures from finite-time {L}yapunov exponents in
  two-dimensional aperiodic flows,'' {\em Physica D: Nonlinear Phenomena},
  vol.~212, pp.~271--304, Dec. 2005.

\bibitem{BruntonRowley_FastComputationFTLE_2010}
S.~L. Brunton and C.~W. Rowley, ``Fast computation of finite-time {L}yapunov
  exponent fields for unsteady flows,'' {\em Chaos: An Interdisciplinary
  Journal of Nonlinear Science}, vol.~20, p.~017503, Mar. 2010.

\bibitem{Haller.Huhn_LAVD_2016}
G.~Haller, A.~Hadjighasem, M.~Farazmand, and F.~Huhn, ``Defining coherent
  vortices objectively from the vorticity,'' {\em Journal of Fluid Mechanics},
  vol.~795, pp.~136--173, May 2016.

\bibitem{HallerYuan_LCSAndMixing_2000}
G.~Haller and G.~Yuan, ``{L}agrangian coherent structures and mixing in
  two-dimensional turbulence,'' {\em Physica D: Nonlinear Phenomena}, vol.~147,
  pp.~352--370, Dec. 2000.

\bibitem{Haller_DynamicPolarDecompostion_2016}
G.~Haller, ``Dynamic rotation and stretch tensors from a dynamic polar
  decomposition,'' {\em Journal of The Mechanics and Physics of Solids}, 2016.

\bibitem{Liberzon.Zimmer_OpenPIV_2021}
A.~Liberzon, T.~Käufer, A.~Bauer, P.~Vennemann, and E.~Zimmer,
  ``Openpiv/openpiv-python: Openpiv-python v0.23.6,'' 2021.

\bibitem{He.Sun_ResNet_2016}
K.~He, X.~Zhang, S.~Ren, and J.~Sun, ``Deep residual learning for image
  recognition,'' in {\em Proceedings of the IEEE conference on computer vision
  and pattern recognition}, pp.~770--778, 2016.

\bibitem{Kirilov.etal_SAM_2023}
A.~Kirillov, E.~Mintun, N.~Ravi, H.~Mao, C.~Rolland, L.~Gustafson, T.~Xiao,
  S.~Whitehead, A.~C. Berg, W.-Y. Lo, {\em et~al.}, ``Segment anything,'' {\em
  arXiv preprint arXiv:2304.02643}, 2023.

\end{thebibliography}


\end{document}